\documentclass{article}

\usepackage[final,nonatbib]{nips_2017}

\usepackage[utf8]{inputenc} 
\usepackage[T1]{fontenc}    
\usepackage{hyperref}       
\usepackage{url}            
\usepackage{booktabs}       
\usepackage{amsfonts}       
\usepackage{nicefrac}       
\usepackage{microtype}      

\usepackage{graphicx}       
\usepackage{textcomp}       
\usepackage{gensymb}        
\usepackage{multirow}       

\usepackage{authblk}        

\title{End-to-End Deep Learning for Steering Autonomous Vehicles Considering Temporal Dependencies}

\author[1]{
  Hesham M. Eraqi\thanks{heraqi@aucegypt.edu}
}
\author[1]{
  Mohamed N. Moustafa\thanks{m.moustafa@aucegypt.edu}
}
\author[2]{
  Jens Honer\thanks{jens.honer@valeo.com}
}
\affil[1]{Department of Computer Science and Engineering, The American University in Cairo, Egypt}
\affil[2]{Valeo Schalter und Sensoren GmbH, Germany}

\begin{document}

\maketitle


\begin{abstract}
Steering a car through traffic is a complex task that is difficult to cast into algorithms. Therefore, researchers turn to training artificial neural networks from front-facing camera data stream along with the associated steering angles. Nevertheless, most existing solutions consider only the visual camera frames as input, thus ignoring the temporal relationship between frames. In this work, we propose a Convolutional Long Short-Term Memory Recurrent Neural Network (C-LSTM), that is end-to-end trainable, to learn both visual and dynamic temporal dependencies of driving. Additionally, We introduce posing the steering angle regression problem as classification while imposing a spatial relationship between the output layer neurons. Such method is based on learning a sinusoidal function that encodes steering angles. To train and validate our proposed methods, we used the publicly available Comma.ai dataset. Our solution improved steering root mean square error by 35\% over recent methods, and led to a more stable steering by 87\%.
\end{abstract}


\section{Introduction} \label{sec:intro}

The 2015 Global Status Report of the World Health Organization (WHO) reported an estimated 1.25 million deaths yearly due to road traffic worldwide \cite{world2014global}. With approximately 89\% of accidents being due to human errors, autonomous vehicles will play a vital role to significantly reduce this number and ultimately to save human lives. With the ability to shift the task of explicitly formulating rules to designing a system that is able to learn those rules, Artificial Intelligence (AI) will play an important role to realize this vision \cite{eraqi2016reactive}. Autonomous vehicles will provide greater mobility for old and disabled people and could reduce energy consumption in transportation by as much as 90\% \cite{brown2013autonomous}. Also, it will translate into less traffic congestion and associated air pollution \cite{brown2013autonomous}.

Lateral control of a vehicle is one of the most fundamental tasks in the design of algorithms to control autonomous vehicles. A human can achieve this almost solely based on visual clues. In AI terms, the human acts as an end-to-end system, i.e. intermediate representations of a driving lane, a planned trajectory or the relation between hand coordination and steering direction may, and probably do, exist somewhere within the brain of the driver but are never explicitly formulated in terms of interfaces.

To understand and recreate the ability to steer a vehicle based on visual clues within an AI system is thus of fundamental importance. To the authors knowledge, the first work in this regard was done by Pomerleau \cite{pomerleau1989alvinn} in 1989 that used a multilayer perceptron (MLP). Since that work, the computational power dramatically increased thanks to massive parallelization in modern GPU’s. In combination with modern network architectures and concepts like Convolutional Neural Networks (CNN) \cite{krizhevsky2012imagenet} that wasn’t known back then. Those advances allowed for having methods that use CNN for end-to-end learning of autonomous vehicles steering \cite{bojarski2016end}, \cite{janai2017computer}, \cite{bojarski2017explaining} that are based on classifying driving scene frame by frame. By design such methods are not equipped to incorporate the temporal relation between image frames and hence cannot learn motion features.

Recurrent Neural Networks (RNN’s) \cite{williams1989learning}, \cite{rodriguez1999recurrent}, \cite{lyu2016learning} represent a class of artificial neural networks that uses memory cells to model the temporal relation between input data and hence learn the underlying dynamics. With the introduction of so called Long Short-Term Memory (LSTM) \cite{hochreiter1997long}, i.e. the ability to remember selectively, modeling long-term relationships became possible within RNN’s.

In this paper we demonstrate quantitatively that a Convolutional Long Short-Term Memory Recurrent Neural Networks (C-LSTM) can significantly improve end-to-end learning performance in autonomous vehicle steering based on camera images. Inspired by the adequacy of CNN in visual feature extraction and the efficiency of Long Short-Term Memory (LSTM) Recurrent Neural Networks in dealing with long-range temporal dependencies our approach allows to model dynamic temporal dependences in the context of steering angle estimation based on camera input. 

Posing regression problems as deep classification problems often shows improvements over direct regression training of CNN’s \cite{rothe2015dex}. We argue that it can still be further improved. Classification tasks assume independence between the output neurons that encode the different classes. However, this assumption loses validity if the classification is used to model a regression. Because, in such case, classes neurons that are spatially close to each other should infer convergent decisions. Here, we propose a method to introduce correlation between the class neurons and thus bridge the gap between full classification problems and regression problems.


\section{System Overview} \label{sec:system_overview}

\begin{figure}[b]
  \centering
  \includegraphics[width=0.8\textwidth]{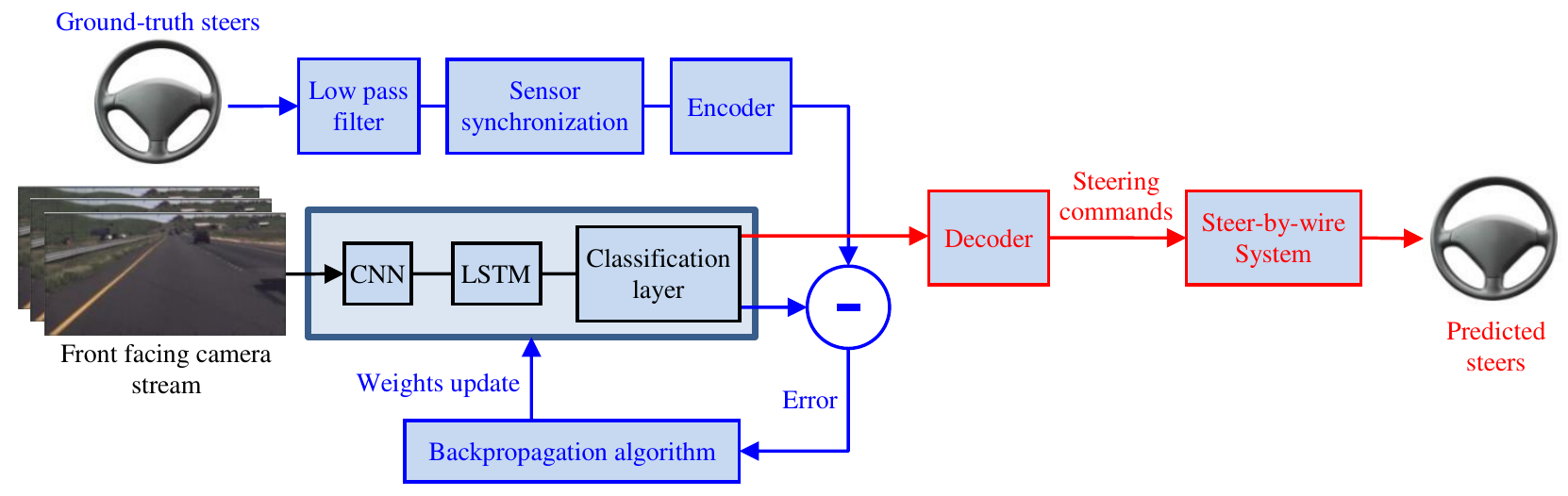}
  \caption{Simplified block diagram of the proposed system. A front-facing RGB camera is used to capture the driving scene and ground-truth steering wheel angles due to a human driver are recorded. During the system deployment phase, a decoder is used to translate the classification layer activations into a steering angle. The blue and red paths represent training and deployment phases respectively.}
  \label{fig:systemBlockDiagram}
\end{figure}

The system we are investigating is comprised of a front-facing RGB camera and a composite neural network consisting of a CNN and LSTM network that estimate the steering wheel angle based on the camera input. Camera images are processed frame by frame by the CNN. The resulting features are then processed within the LSTM network to learn temporal dependences as detailed in section \ref{sec:lstm_arch}. The steering angle prediction is calculated via the output classification layer after the LSTM layers. For full deployment, the steering angle prediction is transmitted via a steer-by-wire system to appropriate actuators.

In the training phase, this setup is extended to include the ground-truth steering angle input together with a filtering mechanism and an encoder, as well as a Backpropagation algorithm \cite{rumelhart1988learning}, \cite{lecun1989backpropagation} to perform the actual training (see figure \ref{fig:systemBlockDiagram}). The details of the training process are described in section \ref{sec:classification_layer}. Figure \ref{fig:systemBlockDiagram} shows a simplified block diagram of the proposed system during training and deployment phases.

Steering wheel sensors are not synchronized with the camera sensor. Typically, the steering wheel angle information is communicated via the vehicle Controller Area Network (CAN) at rates of 100Hz, i.e. at a significantly higher rate than the frame rate per second (FPS) of a camera. We use a low pass filter to reduce the steering wheel sensor measurement noise. Then the smoothed signal is sampled to associate a ground-truth steering angle to each single camera frame. 


\section{C-LSTM Architecture} \label{sec:lstm_arch}

\begin{figure}[ht]
  \centering
  \includegraphics[height=0.4\textwidth]{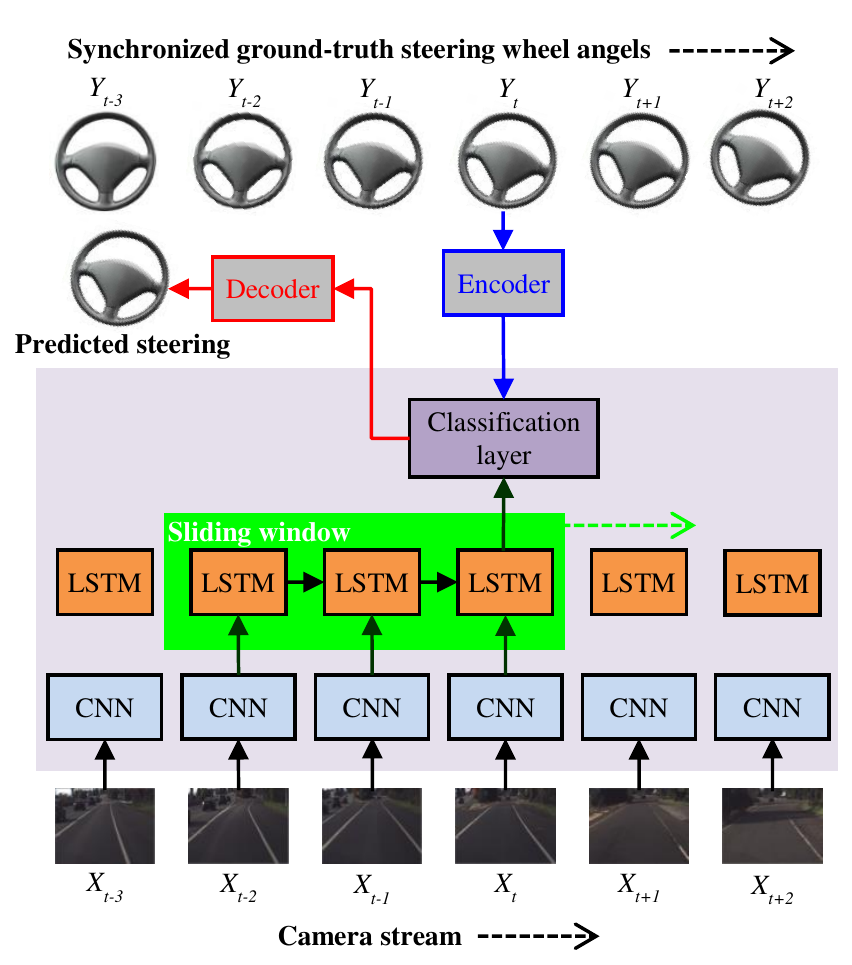}
  \caption{An overview of the proposed C-LSTM architecture unrolled across time. A deep CNN learns to extract best driving scene features, and then the resultant sequential feature vectors are passed into a stack of LSTM layers. The sliding window allows the same frame $X_i$ to be used to train different ground-truth steering angles $Y_i$, but at different states of the LSTM layers. Classification layer is connected via the blue and red paths during the system training and deployment phases respectively.}
  \label{fig:lstmDiagram}
\end{figure}

The proposed Convolutional Long Short-Term Memory Recurrent Neural Network (C-LSTM) architecture combines a deep CNN hierarchical visual feature extractor with a model that can learn to recognize long-term temporal dynamics. Figure \ref{fig:lstmDiagram} depicts the temporal input sequence as processed in our C-LSTM. At each timestamp $t$ the input frame ${X_{t}}$ is processed in a CNN based on the C-LSTM architecture.

CNN’s have been recently applied for large scale recognition problems. The training of deeper CNN architectures is proven to be very successful in such problems \cite{krizhevsky2012imagenet}, \cite{szegedy2015going}, \cite{russakovsky2015imagenet}. This motivated us to choose our CNN to follow the architecture of deepest state-of-the-art CNN’s, and we apply the concept of transfer learning \cite{bengio2012deep}, \cite{donahue2014decaf}. The CNN is pre-trained \cite{bengio2012deep}, \cite{donahue2014decaf} on the Imagenet dataset \cite{deng2009imagenet}, \cite{russakovsky2015imagenet} that features 1.2 million images of approximately 1000 different classes and allows for recognition of a generic set of features and a variety of objects with a high precision. Then, we transfer the trained neural network from that broad domain to another specific one focusing on driving scene images.

The LSTM then processes a sequence of w fixed-length feature vectors (sliding window) from the CNN as depicted in figure \ref{fig:lstmDiagram}. In turn, the LSTM layers learn to recognize temporal dependences leading to a steering decision ${Y_{t}}$ based on the inputs from ${X_{t-w}}$ to ${X_{t}}$. Small values of $t$ lead to faster reactions, but the network learns only short-term dependences and susceptibility for individually misclassified frames increases. Whereas large values of $t$ lead to a smoother behavior, and hence more stable steering predictions, but increase the chance of learning wrong long-term dependences.

In contrast to typical RNN training that is based on fixed subsequent batches, the sliding window concept allows the network to learn to recognize different steering angles from the same frame $X_i$ but at different temporal states of the LSTM layers. Both the CNN and LSTM weights are shared across different steps within the sliding window and in principle this allows for arbitrarily long window size $w$. 

As detailed in section \ref{sec:classification_layer}, we pose the steering angle regression as a classification problem. This is why the single number representing the steering angle $Y_t$ is encoded to a vector of classification layer neurons’ activations. A fully-connected layer with $tanh$ activations is used for the classification layer.

For the domain-specific training, the classification layer of the CNN is re-initialized and trained on camera road data. Training of the LSTM layer is conducted in a many-to-one fashion; the network learns the steering decisions that are associated with intervals of driving. It is important to set the learning rates of each layer appropriately during learning with the Backpropagation algorithm. The classification layer and the LSTM layers use a larger learning rate because it has been initialized with random values. Note that both the CNN and the LSTM are trained jointly at the same time.


\section{Classification Layer} \label{sec:classification_layer}

Autonomous vehicles steering prediction is a continuous value regression problem. Given sensor readings, the system predicts the steering angle as a single continuous value. Neural networks, including CNN, can learn regression problems by using a signal neuron at the output classification layer for the regressed steering angle. This direct approach is adopted by the recent works that tackle this problem in literature \cite{bojarski2016end}, \cite{janai2017computer}, \cite{bojarski2017explaining}.

In \cite{rothe2015dex}, regression formulation through a deep classification followed by expected value refinement has been introduced to improve accuracy of the apparent age estimation from images. The regression problem is posed as a deep classification problem followed by a $softmax$ expected value refinement. We argue that such approach can still be further improved mainly because it has two downsides.

\begin{figure}[ht]
  \centering
  \includegraphics[width=1.0\textwidth]{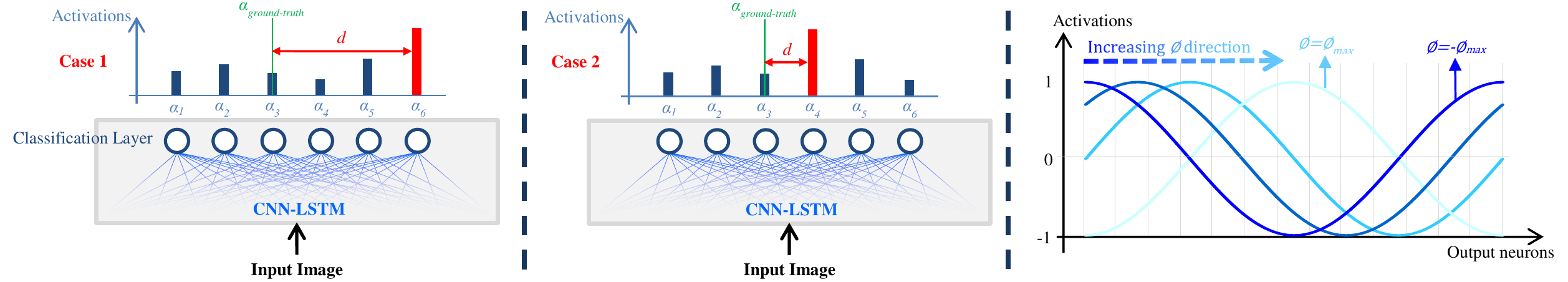}
  \caption{In case 1, low activations occur near the ground-truth neuron, whereas the neuron with peak activation is situated on the far right side, i.e. far away from the ground truth. In contrast, case 2 features the neuron with peak activation close to the neuron that is expected by the ground-truth and thus a better prediction. Yet loss metrics that neglect the spatial relations will yield the same penalties in both cases. The figure also shows Sine waves encoding at different steering angles. The dotted arrow denotes the increasing $\phi$ direction.}
  \label{fig:nllCases_shifted_sins}
\end{figure}

The first downside is that it neglects the topological relation between output classes, i.e. the network has no notion of how different two steering angles are. Figure \ref{fig:nllCases_shifted_sins} emphasizes the topological problem. With the steering angles being mapped to discrete output neurons, each output neuron $i$ spans a small steering range of ${\alpha_{i}}$. The figure depicts two prediction example cases that both yield the same loss for conventional loss metrics like Negative Log Likelihood (NLL) or mean squared error (MSE).

The second drawback of the conventional formulation of regression as a classification problem is that the amount of required training data scales with the number of discrete angles, i.e. classes. The expanding number of class labels requires collecting more training data such that all class labels contain sufficient training volumes. Conversely, if the training data is limited, so is the number of classes.

To solve this problem we introduce a spatial relationship between the classification layer neurons. The method is based on learning an arbitrary function that encodes the steering angle. It's inspired from our patent regarding pose angle estimation of an object in an input image \cite{moustafa2005system}. The steering wheel sensor provides a steering angle ${\phi}$ ranging from ${-\phi_{max}^\circ}$ (extreme turning to the left) to ${\phi_{max}^\circ}$ (extreme right turn). ${\phi=0^\circ}$ means the vehicle driving straight forward. We choose the encoding function to be a sine wave and the steering angle to be its phase shift as in (\ref{eq:phaseshift}):
\begin{equation} \label{eq:phaseshift}
Y_i = \sin \left( {\frac{2\pi (i-1)}{N-1} - \frac{\phi\pi}{2\phi_{max}}} \right) , \,\,\,\,\, 1 \leq i \leq N ,
\end{equation}
where $Y_i$ is the activation of the output neuron i and N is the number of output layer neurons, i.e. classes. Such choice for the learned function and its parameter(s) encoding steering, to be a sine wave and its phase shift, guarantees gradual changes in steering to cause gradual changes of the output layer activations. Figure \ref{fig:nllCases_shifted_sins} shows example sine waves encoding different steering wheel angles.

\begin{figure}[ht]
  \centering
  \includegraphics[height=0.3\textwidth]{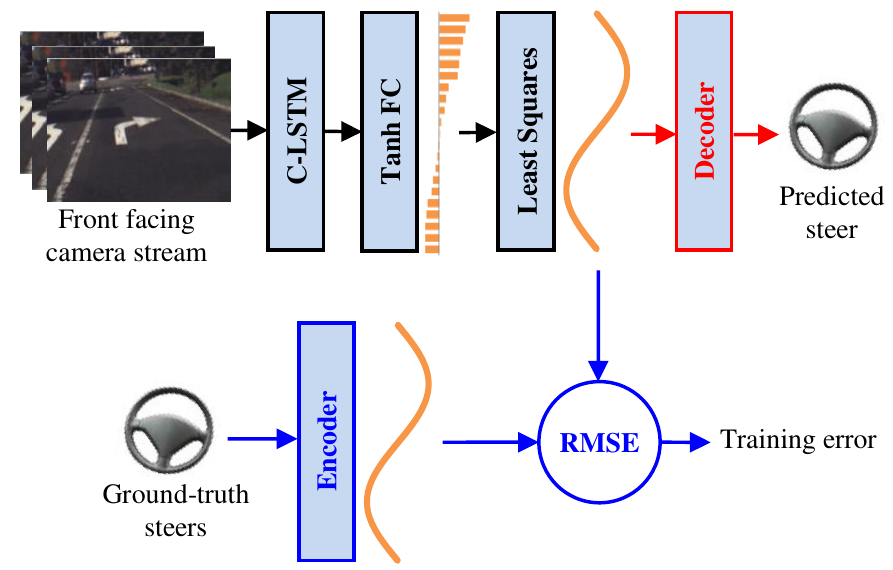}
  \caption{Fitting a sin wave that encodes driving steering wheel angle. The blue and red paths represent training and deployment phases respectively.}
  \label{fig:rmse_calc}
\end{figure}

Figure \ref{fig:rmse_calc} shows how the proposed method is used for learning efficient vehicle steering. $Tanh$ activation functions are used for the classification layer neurons to allow it shaping sine waves with amplitude from -1 to 1. During training, the ground-truth steering angle $\phi$ is encoded as a sine wave function as in (\ref{eq:phaseshift}). On the other hand, the classification layer output uses a Least Squares regression in order to fit the predicted function. The Backpropagation loss function is calculated as the root mean square error (RMSE) between the two waveforms over batches of sequential data.

During deployment the steering angle is decoded from the Least Squares regression result. The decoder calculates the sine wave phase shift that contains the information about the steering angle. This new concept of network output formulation as a fitted sinusoidal function can be extended to other application domains where regression is posed as a classification problem for better performance; like Face Age Detection, Face Pose Estimation, House Price Estimation, etc.

\section{Experimental Results} \label{sec:results}

In this section, we first introduce the dataset we used for experimental work and the system evaluation metrics. Then we present the implementation details of our method, describe experimental setups, and discuss results.


\subsection{Dataset and Evaluation Metrics} \label{sec:results_metrics}

We trained and validated our system on the recently publicly released database by “comma.ai” in \cite{santana2016learning}. It contains 7.26 hours of highway and city driving data that are divided in 11 videos captured during both day and night. The dataset also has several sensors that were measured in different frequencies. In this work, we only use the camera frames and steering wheel angle signal. In \cite{bojarski2016end}, the authors trained their system using about 72 hours of driving data that is not publicly available. We aim to provide a baseline performance for future works to benchmark against.

We choose the first and third video files, having 1 hour of driving, to form the testing set, and the remaining nine files were kept for training and validation. That balances the training and testing sets between day and night driving. We sampled both of the training and testing datasets to 2 Hz. In \cite{xu2016end}, Comma.ai dataset is used to validate Deep Predictive coding networks. In that work, random 5\% chunks are sampled from each video file in the dataset and kept for validation and testing. Such split strategy causes the correlation between the training and testing data to be too high, and the steering prediction becomes a much easier problem. We separate complete video files for testing such that testing data contains driving sessions that are completely different from those of the training data, i.e.; different locations, time of day, and driving conditions.

Both Mean Absolute Error (MAE) and Root Mean Square Error (RMSE) can express average system prediction error. For steering angle prediction problem, large errors are particularly undesirable. We choose to report our results using RMSE, defined as in (\ref{eq:rmse}) to gives a relatively high weight to large errors.
\begin{equation} \label{eq:rmse}
RMSE = \sqrt {\frac{1}{|D|} \sum_{i=1}^{|D|} \left( G_i-P_i \right)^2 } ,
\end{equation}
where $G_i$ and $P_i$ are the ground-truth and predicted steering angles for a frame $i$ respectively in a testing set having a total of $|D|$ frames. Both angles are measured in angular degrees.

MAE and RMSE can range from $0$ to $\inf$ and are indifferent to the direction of errors. They cannot infer how much stable steering is. Thus, we use another metric $W$ that measures the whiteness of the predicted steering signal as in (\ref{eq:whitness}). The smaller the value of $W$, the more smooth steering variations are predicted.
\begin{equation} \label{eq:whitness}
W = {\frac{1}{|D|} \sum_{i=1}^{|D|} \left( \left. \frac{\partial P(t)}{\partial t} \right|_{t=i} \right)^2 }
\end{equation}


\subsection{Results} \label{sec:results_results}

We report in this section our experimental results on the testing dataset we defined earlier. In table \ref{table:CNNsResults}, we report performance of using a CNN in learning steering using regression; the CNN has a single output neuron. Firstly, we use a simple CNN that follows the architecture introduced in \cite{bojarski2016end}. And then we apply transfer learning using deeper state of the art networks, which are the third version of Inception network \cite{szegedy2016rethinking} and Resnet network having 152 layers \cite{he2016deep}.

\begin{table}[h]
  \caption{Regression using CNN}
  \label{table:CNNsResults}
  \centering
  \def\arraystretch{1.5}
  
  \begin{tabular}{lcc}
  \hline
  \multicolumn{1}{c}{\multirow{2}{*}{CNN Network}}
  & \multicolumn{2}{c}{Performance}
  \\ \cline{2-3} 
  \multicolumn{1}{c}{}
  & \begin{tabular}[c]{@{}c@{}}RMSE {[} Degrees {]}\end{tabular}
  & \begin{tabular}[c]{@{}c@{}}Whiteness {[} Degrees / Time unit {]}\end{tabular}
  \\ \hline
  Simple CNN \cite{bojarski2016end} & 23.30 & 65.8
  \\ \hline
  Inception V3 & 18.67 & 43.9
  \\ \hline
  Resnet 152 & 17.77 & 39.1
  \\ \hline
  \end{tabular}
\end{table}

Table \ref{table:CNNsResults} confirms that deeper networks perform better. We conducted cross-validation using different optimizers, network regulation, and learning rates until we found best choices. The simple CNN and the fully connected layers had a learning rate of $10^{-3}$, while the convolutional layers had a learning rate of $10^{-5}$ for transfer learning . Adam optimization \cite{kingma2014adam} was used to train all networks. Applying batch normalization and dropout techniques helped minimizing over-fitting.

Table \ref{table:CNNvsLSTM} compares the direct regression performance with the state-of-the-art classification method in \cite{rothe2015dex}. Subsequently, it compares them with our proposed classification method; by fitting a sinusoidal function. Most importantly, it reports the results of our complete solution which uses the C-LSTM architecture. The output feature vector from the CNN is of length 2048. Empirically, $\theta$, $\phi_{max}$, and $\sigma^2$ are chosen to be equal to $4^\circ$, $190^\circ$, and $80$ respectively. Hence, the number of output neurons $N$ is equal to $95$ neurons. We found that a sliding window stride of $1$ time step resulted into most accurate predictions. Empirically, we used $2$ LSTM layers, with each layer having $500$ neurons and $w$ covering $5$ seconds of driving.

\begin{table}[h]
  \caption{Comparison of CNN And C-LSTM For Regression and Classification}
  \label{table:CNNvsLSTM}
  \centering
  \resizebox{\textwidth}{!}{
  \def\arraystretch{1.5}
  \begin{tabular}{lcccc}
  \hline
  \multicolumn{1}{c}{\multirow{2}{*}{CNN Network}} & \multicolumn{2}{c}{CNN Performance}
  & \multicolumn{2}{c}{C-LSTM Performance}
  \\ \cline{2-5} 
  \multicolumn{1}{c}{} & \begin{tabular}[c]{@{}c@{}}RMSE {[} Degrees {]}\end{tabular}
  & \begin{tabular}[c]{@{}c@{}}Whiteness {[} Degrees / Time unit {]}\end{tabular}
  & \begin{tabular}[c]{@{}c@{}}RMSE {[} Degrees {]}\end{tabular}
  & \begin{tabular}[c]{@{}c@{}}Whiteness {[} Degrees / Time unit {]}\end{tabular}
  \\ \hline
  Regression & 17.77 & 39.1 & 16.01 & 9.7
  \\ \hline
  Classification,using NLL \cite{rothe2015dex} & 18.70 & 54.1 & 17.84 & 10.0  
  \\ \hline
  Classification by,sine wave fitting & 17.44 & 43.9 & \textbf{14.93} & \textbf{8.2}
  \\ \hline
  \end{tabular}
  }
\end{table}

Table \ref{table:CNNvsLSTM} demonstrates that our proposed method for classification performed better than both of the conventional classification method and the direct regression of steering angle method. Most importantly, it shows that using our C-LSTM architecture led to a significant improvement of steering angle prediction in terms of accuracy and stability reflected by the predicted steering signal RMSE and whiteness respectively. The signal was more accurate than the state-of-the-art solution by 35\% and was more stable by 87\%. A steering wheel angle RMSE slightly less than $20^\circ$ is still an achievement, since modern car ``steering ratio'' is in the range $12:1$ to $20:1$ \cite{pacejka2005tire}. Consequently, a steering wheel RMSE of $20^\circ$ is actually equivalent to approximately only $1^\circ$ of vehicle wheels (tyres) off-direction.


\section{Conclusions} \label{sec:conclusion}

We propose and benchmark a C-LSTM architecture that allows learning both visual and temporal dependencies of driving. We also introduce posing the steering angle regression problem as a deep classification problem by imposing a spatial relationship between the output layer neurons. That concept of output formulation as a fitted sinusoidal function can be extended to other application domains where regression is posed as a classification problem for better performance. Our solution achieved 35\% steering RMSE improvement and led to a more stable steering by 87\% compared to recent methods tested on the benchmark Comma.ai publicly available dataset. The future work should focus on verifying the proposed method via simulation and analyzing the effect of the number of output neurons on the steering performance.


\small
\bibliographystyle{plain}
\bibliography{references.bib}

\begin{thebibliography}{10}

\bibitem{bengio2012deep}
Yoshua Bengio et~al.
\newblock Deep learning of representations for unsupervised and transfer
  learning.
\newblock {\em ICML Unsupervised and Transfer Learning}, 27:17--36, 2012.

\bibitem{bojarski2016end}
Mariusz Bojarski, Davide Del~Testa, Daniel Dworakowski, Bernhard Firner, Beat
  Flepp, Prasoon Goyal, Lawrence~D Jackel, Mathew Monfort, Urs Muller, Jiakai
  Zhang, et~al.
\newblock End to end learning for self-driving cars.
\newblock {\em arXiv preprint arXiv:1604.07316}, 2016.

\bibitem{bojarski2017explaining}
Mariusz Bojarski, Philip Yeres, Anna Choromanska, Krzysztof Choromanski,
  Bernhard Firner, Lawrence Jackel, and Urs Muller.
\newblock Explaining how a deep neural network trained with end-to-end learning
  steers a car.
\newblock {\em arXiv preprint arXiv:1704.07911}, 2017.

\bibitem{brown2013autonomous}
Austin Brown, Brittany Repac, and Jeff Gonder.
\newblock Autonomous vehicles have a wide range of possible energy impacts.
\newblock Technical report, NREL, University of Maryland, 2013.

\bibitem{deng2009imagenet}
Jia Deng, Wei Dong, Richard Socher, Li-Jia Li, Kai Li, and Li~Fei-Fei.
\newblock Imagenet: A large-scale hierarchical image database.
\newblock In {\em Computer Vision and Pattern Recognition, 2009. CVPR 2009.
  IEEE Conference on}, pages 248--255. IEEE, 2009.

\bibitem{donahue2014decaf}
Jeff Donahue, Yangqing Jia, Oriol Vinyals, Judy Hoffman, Ning Zhang, Eric
  Tzeng, and Trevor Darrell.
\newblock Decaf: A deep convolutional activation feature for generic visual
  recognition.
\newblock In {\em Icml}, volume~32, pages 647--655, 2014.

\bibitem{eraqi2016reactive}
Hesham Eraqi, Youssef EmadEldin, and Mohamed Moustafa.
\newblock Reactive collision avoidance using evolutionary neural networks.
\newblock In {\em Proceedings of the 8th International Joint Conference on
  Computational Intelligence}, volume~1, 2016.

\bibitem{he2016deep}
Kaiming He, Xiangyu Zhang, Shaoqing Ren, and Jian Sun.
\newblock Deep residual learning for image recognition.
\newblock In {\em Proceedings of the IEEE Conference on Computer Vision and
  Pattern Recognition}, pages 770--778, 2016.

\bibitem{hochreiter1997long}
Sepp Hochreiter and J{\"u}rgen Schmidhuber.
\newblock Long short-term memory.
\newblock {\em Neural computation}, 9(8):1735--1780, 1997.

\bibitem{janai2017computer}
Joel Janai, Fatma G{\"u}ney, Aseem Behl, and Andreas Geiger.
\newblock Computer vision for autonomous vehicles: Problems, datasets and
  state-of-the-art.
\newblock {\em arXiv preprint arXiv:1704.05519}, 2017.

\bibitem{kingma2014adam}
Diederik Kingma and Jimmy Ba.
\newblock Adam: A method for stochastic optimization.
\newblock {\em arXiv preprint arXiv:1412.6980}, 2014.

\bibitem{krizhevsky2012imagenet}
Alex Krizhevsky, Ilya Sutskever, and Geoffrey~E Hinton.
\newblock Imagenet classification with deep convolutional neural networks.
\newblock In {\em Advances in neural information processing systems}, pages
  1097--1105, 2012.

\bibitem{lecun1989backpropagation}
Yann LeCun, Bernhard Boser, John~S Denker, Donnie Henderson, Richard~E Howard,
  Wayne Hubbard, and Lawrence~D Jackel.
\newblock Backpropagation applied to handwritten zip code recognition.
\newblock {\em Neural computation}, 1(4):541--551, 1989.

\bibitem{lyu2016learning}
Haobo Lyu, Hui Lu, and Lichao Mou.
\newblock Learning a transferable change rule from a recurrent neural network
  for land cover change detection.
\newblock {\em Remote Sensing}, 8(6):506, 2016.

\bibitem{moustafa2005system}
Mohamed~Nabil Moustafa.
\newblock System and method for pose-angle estimation, October~25 2005.
\newblock US Patent 6,959,109.

\bibitem{world2014global}
World~Health Organization.
\newblock {\em Global status report on alcohol and health 2014}.
\newblock World Health Organization, 2014.

\bibitem{pacejka2005tire}
Hans Pacejka.
\newblock {\em Tire and vehicle dynamics}.
\newblock Elsevier, 2005.

\bibitem{pomerleau1989alvinn}
Dean~A Pomerleau.
\newblock Alvinn, an autonomous land vehicle in a neural network.
\newblock Technical report, Carnegie Mellon University, Computer Science
  Department, 1989.

\bibitem{rodriguez1999recurrent}
Paul Rodriguez, Janet Wiles, and Jeffrey~L Elman.
\newblock A recurrent neural network that learns to count.
\newblock {\em Connection Science}, 11(1):5--40, 1999.

\bibitem{rothe2015dex}
Rasmus Rothe, Radu Timofte, and Luc Van~Gool.
\newblock Dex: Deep expectation of apparent age from a single image.
\newblock In {\em Proceedings of the IEEE International Conference on Computer
  Vision Workshops}, pages 10--15, 2015.

\bibitem{rumelhart1988learning}
David~E Rumelhart, Geoffrey~E Hinton, and Ronald~J Williams.
\newblock Learning representations by back-propagating errors.
\newblock {\em Cognitive modeling}, 5(3):1, 1988.

\bibitem{russakovsky2015imagenet}
Olga Russakovsky, Jia Deng, Hao Su, Jonathan Krause, Sanjeev Satheesh, Sean Ma,
  Zhiheng Huang, Andrej Karpathy, Aditya Khosla, Michael Bernstein, et~al.
\newblock Imagenet large scale visual recognition challenge.
\newblock {\em International Journal of Computer Vision}, 115(3):211--252,
  2015.

\bibitem{santana2016learning}
Eder Santana and George Hotz.
\newblock Learning a driving simulator.
\newblock {\em arXiv preprint arXiv:1608.01230}, 2016.

\bibitem{szegedy2015going}
Christian Szegedy, Wei Liu, Yangqing Jia, Pierre Sermanet, Scott Reed, Dragomir
  Anguelov, Dumitru Erhan, Vincent Vanhoucke, and Andrew Rabinovich.
\newblock Going deeper with convolutions.
\newblock In {\em Proceedings of the IEEE Conference on Computer Vision and
  Pattern Recognition}, pages 1--9, 2015.

\bibitem{szegedy2016rethinking}
Christian Szegedy, Vincent Vanhoucke, Sergey Ioffe, Jon Shlens, and Zbigniew
  Wojna.
\newblock Rethinking the inception architecture for computer vision.
\newblock In {\em Proceedings of the IEEE Conference on Computer Vision and
  Pattern Recognition}, pages 2818--2826, 2016.

\bibitem{williams1989learning}
Ronald~J Williams and David Zipser.
\newblock A learning algorithm for continually running fully recurrent neural
  networks.
\newblock {\em Neural computation}, 1(2):270--280, 1989.

\bibitem{xu2016end}
Huazhe Xu, Yang Gao, Fisher Yu, and Trevor Darrell.
\newblock End-to-end learning of driving models from large-scale video
  datasets.
\newblock {\em arXiv preprint arXiv:1612.01079}, 2016.

\end{thebibliography}

\end{document}